\documentclass[conference]{IEEEtran}
\IEEEoverridecommandlockouts
\usepackage{cite}
\usepackage{amsmath,amssymb,amsfonts}
\usepackage{algorithmic}
\usepackage{booktabs} 
\usepackage{adjustbox} 
\usepackage{caption} 
\usepackage{array} 
\usepackage{graphicx}
\usepackage{textcomp}
\usepackage{subcaption}
\usepackage{xcolor}
\renewcommand{\arraystretch}{0.8} 

\def\BibTeX{{\rm B\kern-.05em{\sc i\kern-.025em b}\kern-.08em
    T\kern-.1667em\lower.7ex\hbox{E}\kern-.125emX}}
\begin{document}

\title{Graph Pruning Based Spatial and Temporal Graph Convolutional Network with Transfer Learning for Traffic Prediction \\
}

\author{
    \IEEEauthorblockN{Zihao Jing}
    \IEEEauthorblockA{\textit{Department of Computer Science} \\
    \textit{Western University}\\
    zjing29@uwo.ca}
    \\
    \IEEEauthorblockN{Ganlin Feng}
    \IEEEauthorblockA{\textit{Department of Computer Science} \\
    \textit{Western University}\\
    gfeng23@uwo.ca}
    \and
    \IEEEauthorblockN{Yuxi Long}
    \IEEEauthorblockA{\textit{Department of Computer Science} \\
    \textit{Western University}\\
    ylong66@uwo.ca}
    \\
    \IEEEauthorblockN{Morgan He}
    \IEEEauthorblockA{\textit{Department of Computer Science} \\
    \textit{Western University}\\
    mhe52@uwo.ca}
}

\maketitle

\begin{abstract}
With the process of urbanization and the rapid growth of population, the issue of traffic congestion has become an increasingly critical concern. Intelligent transportation systems heavily rely on real-time and precise prediction algorithms to address this problem. While Recurrent Neural Network (RNN) and Graph Convolutional Network (GCN) methods in deep learning have demonstrated high accuracy in predicting road conditions when sufficient data is available, forecasting in road networks with limited data remains a challenging task. This study proposed a novel Spatial-temporal Convolutional Network (TL-GPSTGN) based on graph pruning and transfer learning framework to tackle this issue. Firstly, the essential structure and information of the graph are extracted by analyzing the correlation and information entropy of the road network structure and feature data. By utilizing graph pruning techniques, the adjacency matrix of the graph and the input feature data are processed, resulting in a significant improvement in the model's migration performance. Subsequently, the well-characterized data are inputted into the spatial-temporal graph convolutional network to capture the spatial-temporal relationships and make predictions regarding the road conditions. Furthermore, this study conducts comprehensive testing and validation of the TL-GPSTGN method on real datasets, comparing its prediction performance against other commonly used models under identical conditions. The results demonstrate the exceptional predictive accuracy of TL-GPSTGN on a single dataset, as well as its robust migration performance across different datasets.
\end{abstract}


\section{Introduction}
Transportation plays a critical role in modern society, serving as the backbone of daily activities and economic interactions. According to a survey in 2019, drivers undertook an average of 2.22 daily driving trips during 2016-2017, covering approximately 31.5 miles per day \cite{b1}. However, the rapid acceleration of urbanization and population growth has exacerbated urban traffic congestion, turning it into a pressing global challenge. Traffic congestion not only hinders the daily travel of urban residents but also imposes significant economic, social, and environmental costs. From wasted fuel and lost productivity to increased air pollution and greenhouse gas emissions, the consequences of congestion ripple far beyond the roads themselves. As congestion is both periodic and episodic in nature, it remains difficult to detect and mitigate promptly, emphasizing the urgent need for algorithms that can predict traffic conditions accurately and efficiently.

Traffic condition prediction leverages advancements in big data and artificial intelligence to forecast road conditions at each node in a road network by analyzing historical and real-time traffic data (e.g., camera images, electronic transactions, and license plate identification data) \cite{b2}. By providing forward-looking insights, these predictions assist traffic managers in making informed decisions to improve road conditions and alleviate congestion. Beyond management, accurate predictions can also enable smart city initiatives, optimize logistics, and enhance public transportation systems.

Over the past few decades, breakthroughs in technologies such as the Internet of Things (IoT), Big Data, and Artificial Intelligence have led to the development of various traffic prediction models, including Regression \cite{b3}, Long Short-Term Memory (LSTM) \cite{b4}, and Graph Convolutional Networks (GCN) \cite{b5}. These models have demonstrated high performance on specific datasets, making significant strides in traffic forecasting. However, they are often limited to the regions for which they were trained, as differences in road network structures, data availability, and informatization levels across regions hinder their generalizability. Many regions lack sufficient high-quality training data, making it challenging to apply existing models effectively to new road networks \cite{b6}. This highlights an urgent need for models with strong transferability that can adapt to diverse traffic scenarios.

To address the limitations of existing traffic prediction models, this paper explores the application of transfer learning techniques in traffic prediction. Transfer learning leverages knowledge from well-studied regions with abundant data to improve predictions in regions with sparse data. Building on this foundation, this paper introduces a novel graph pruning feature filtering method to enhance the migration performance of traffic prediction models. By integrating graph pruning with a spatial-temporal graph convolutional network, we propose the TL-GPSTGN model— a spatial-temporal graph convolutional network based on transfer learning and graph pruning.

The TL-GPSTGN model addresses the challenges of data scarcity and transferability, enabling accurate predictions in new road networks with minimal training data. A series of experiments demonstrate that TL-GPSTGN achieves superior prediction accuracy compared to existing models while excelling in migration performance across different road networks. By making the rapid deployment of prediction models to new road networks feasible, TL-GPSTGN represents a significant step toward the practical implementation of intelligent transportation systems. This work not only advances the state of traffic prediction research but also contributes to the broader development of smart cities, paving the way for more efficient and sustainable transportation systems.

\section{Background}

\subsection{Prediction Techniques}

The traffic prediction task is an essential component of realizing smart transportation systems. By leveraging advanced algorithms and artificial intelligence (AI) models, traffic prediction aims to forecast future traffic flow and congestion based on historical traffic data, road network structures, and real-time monitoring inputs. Accurate traffic predictions enable traffic managers to make informed decisions, implement effective traffic control measures, and enhance transportation efficiency. Beyond management, traffic prediction also supports smart city initiatives by optimizing public transportation, improving logistics, and reducing environmental impacts caused by traffic congestion.

Over the last few decades, various traffic prediction methods have been developed, which can be broadly categorized into traditional methods and deep learning-based methods. Traditional traffic forecasting methods, often rooted in statistical modeling and time series analysis, include widely used techniques such as the autoregressive moving average model (ARIMA), exponential smoothing, and the gray system model \cite{b7}. These methods rely on identifying trends and periodicities in historical traffic data and applying mathematical models to predict future conditions. While effective in simpler scenarios, traditional methods face significant limitations. They require manual adjustment of numerous model parameters and struggle to capture the complexities of spatio-temporal relationships in dynamic and interconnected traffic systems.

In recent years, the rapid advancement of deep learning has revolutionized traffic prediction, enabling substantial improvements in accuracy and scalability. Deep learning-based models, such as Graph Convolutional Neural Networks (GCN) and Recurrent Neural Networks (RNN), have become powerful tools for handling spatio-temporal data. GCNs excel at modeling spatial relationships in road networks using adjacency matrices to represent node connections, while RNNs specialize in processing sequential data and addressing temporal dependencies. Combining these two approaches, Spatial-Temporal Graph Convolutional Networks (STGCN) integrate spatial and temporal features to predict traffic flow with remarkable accuracy. By modeling the road network structure and incorporating time-series data, STGCNs provide a robust framework for understanding and forecasting traffic flow and congestion \cite{b8}.

Despite these advancements, challenges remain. Deep learning models often require large amounts of high-quality training data, and their interpretability is sometimes limited due to the complexity of their neural network structures. Moreover, the ability to generalize these models across regions with diverse road networks and traffic conditions is an ongoing challenge. Addressing these issues is pivotal for developing practical and scalable traffic prediction solutions.

In summary, traffic prediction methods have evolved significantly, from statistical models to complex deep learning frameworks. With the ongoing development of techniques such as graph convolutional networks and their extensions, the future holds promise for more accurate, efficient, and interpretable traffic prediction models. These advancements will provide better decision support for urban transportation planning, fostering the development of smarter and more sustainable transportation systems.

\subsection{Problem Definition}
The traffic forecasting task involves predicting future traffic flows based on historical traffic data. Accurate road condition prediction results are provided through modeling and analytical reasoning. There should be several monitoring points or sensors in a traffic road network that are capable of generating a stable output of road condition characterization data. A piece of raw data $X_t$ at moment $t$ is an $n$-dimensional vector whose dimension $n$ is the number of sensors in the observed road network. The data generated by each sensor will be continuously mapped to a fixed location in this vector. The sensor will $i$ will record the number of vehicles $x_{t,i}$ that pass through this sensor at time $\Delta t$ , and let the number of vehicles that pass through sensor $i$ at a given instant be $cnt$:
\begin{equation}
    x_{t,i} = \int_t^{t+\Delta t} cnt \, dt
\end{equation}
\begin{equation}
     X_t = (x_{t,1}, x_{t,2}, \dots, x_{t,n})
\end{equation}
\begin{equation}
    D = (X_1, X_2, \dots, X_k)^T
\end{equation}

Since $\Delta t$ time is usually a few minutes, which is very short compared to the observation history and prediction intervals, the road condition data in the network at time t can be represented as a vector $X_t$. A series of road condition data at certain intervals can be represented as a matrix $D$. The matrix $D$ constitutes the set of road condition data for a particular road network.

Correlation adjacency matrix: a matrix that represents the strength of relationships between nodes in a graph. For a graph containing n nodes, its correlation adjacency matrix $A$ is an $n \times n$ matrix, where $A[i][j]$ denotes the correlation between node $i$ and node $j$. The correlation matrix $A[i][j]$ represents the correlation between node $j$ and node $i$.

In the traffic road network, the number of nodes in the correlation adjacency matrix should be consistent with the road data vector $X$, and the corresponding positions of each node should also be uniform.

The input of traffic prediction should contain a correlation adjacency matrix and time series feature data, and the output is the road flow after a specified time. The correlation adjacency matrix is the $n \times n$ symmetric matrix $A$. Assuming that the number of feature data is his and the prediction interval is pred, the input-output relationship for the prediction task is

\begin{equation}
    Y = X_{t+pred} = F(X_{t-his}, X_{t-his+1}, \dots, X_t)
\end{equation}

Through the use of pre-trained models, historical data can be used as input to make predictions about future traffic states. These predictions can provide transportation managers with important references regarding transportation planning, thus supporting more effective urban traffic management and improving traffic conditions.

\subsection{Literature Review}
Traffic prediction has been extensively studied, and various methods have been proposed to tackle its inherent challenges. Raei et al. \cite{rae2024gps} utilized spatial-temporal modeling based on GPS data to predict evacuee behavior during emergencies. Their approach effectively integrated spatial dependencies, allowing for precise predictions in crisis scenarios. However, the inherent sparsity and variability of GPS data across different regions significantly limited its scalability and practical applicability to broader traffic systems.

Yoosuf et al. \cite{yoosuf2024perceptual} proposed a deep learning framework designed to enhance the perceptual quality of spatiotemporal traffic forecasting in sparse data environments. By focusing on improving accuracy through innovative modeling techniques, their method demonstrated promising results. Nevertheless, it underperformed in capturing abrupt and nonlinear changes in traffic dynamics, which are common in real-world urban environments.

Luo et al. \cite{luo2024ad_hoc} explored real-time traffic communication prediction within vehicle ad-hoc networks (VANETs) using machine learning techniques. Their dynamic network models effectively handled communication data variability, but the model's performance degraded in nonlinear and highly dynamic traffic scenarios. This highlighted the need for more resilient methodologies capable of adapting to unpredictable vehicular networks.

Kim et al. \cite{kim2024backbone} developed a Prophet-based variance reduction model to predict traffic patterns in IP backbone networks. Their approach achieved notable accuracy improvements in specific use cases, particularly under stable network conditions. However, the model struggled to generalize to multi-dimensional and interconnected traffic systems, limiting its applicability to complex and dynamic road networks.

Visconti et al. \cite{visconti2024bayesian} applied Bayesian machine learning techniques with spatio-temporal logic to model traffic flow and account for uncertainties. Their probabilistic reasoning framework proved robust for handling noise and variability in traffic data, making it a valuable tool for certain applications. However, the method lacked scalability for large datasets with complex topologies, making it challenging to deploy in expansive urban traffic systems.

Manibardo et al. \cite{manibardo2024forecasting} investigated short-term traffic forecasting in environments with limited data availability, introducing strategies to mitigate data scarcity. While their method was effective on well-structured and low-noise datasets, it showed limitations when applied to more chaotic and unstructured real-world traffic scenarios, highlighting the need for approaches capable of handling variability.

Lim et al. \cite{lim2024airborne} focused on reducing airborne delays in terminal areas using machine learning-powered extended arrival management. By leveraging historical data, their approach effectively improved scheduling efficiency. However, its reliance on high-quality and structured data restricted its adaptability to less-documented or poorly monitored environments.

Zhang et al. \cite{zhang2024cooperative} proposed a reinforcement learning-based cooperative control system to optimize traffic signals and connected automated vehicles. Their approach demonstrated potential for improving traffic safety and efficiency but faced challenges related to computational costs and scalability, limiting its applicability in densely populated urban networks where real-time responsiveness is essential.

Wu et al. \cite{wu2024vessel} developed a time-series data fusion model for vessel trajectory prediction, which provided significant advancements in maritime traffic management. However, the approach showed limited adaptability to other domains, such as urban or road traffic, where dynamic and interconnected systems present additional complexities.

These studies highlight diverse approaches and significant contributions to traffic prediction, ranging from spatial-temporal modeling and deep learning frameworks to Bayesian methods and reinforcement learning. Despite their innovations, these methods often exhibit notable limitations in generalizability, scalability, and adaptability to sparse or dynamic datasets. These challenges underscore the need for more flexible, transferable, and robust methods capable of addressing real-world traffic prediction scenarios.

\section{Methodology}

This section starts with a general overview of the TL-GPSTGN model structure and then describes each of its modules in detail.

\subsection{Model Architecture Overview}
To solve the traffic prediction problem with insufficient training data, this spatial- temporal graph convolutional network-based transfer learning model (TL-GPSTGN) is constructed. The TL-GPSTGN model consists of the modules of graph pruning processor (GPP), spatial-temporal graph convolutional network (STGCN), and Reductor. Among them, GPP consists of Information Entropy Analyzer (IEA), Graph Pruning (GP) module, and Normalization module, which are used to optimize the input spatial-temporal data. STGCN incorporates the ideas of graph convolutional network and time series modeling, which can effectively capture the relationship between time and space. Compared with the traditional convolutional neural network (CNN) and recurrent neural network (RNN), STGCN is able to deal with non-regular and inhomogeneous spatial-temporal data, which is suitable for coping with complex spatial-temporal structures in road networks \cite{b8}. The Reductor module is used to reconstruct the output of the model into traffic flow, i.e., the prediction result, based on the normalization rules. The structure of the model is shown in Figure 1.

\begin{figure}[htbp]
\centerline{\includegraphics[width=\linewidth]{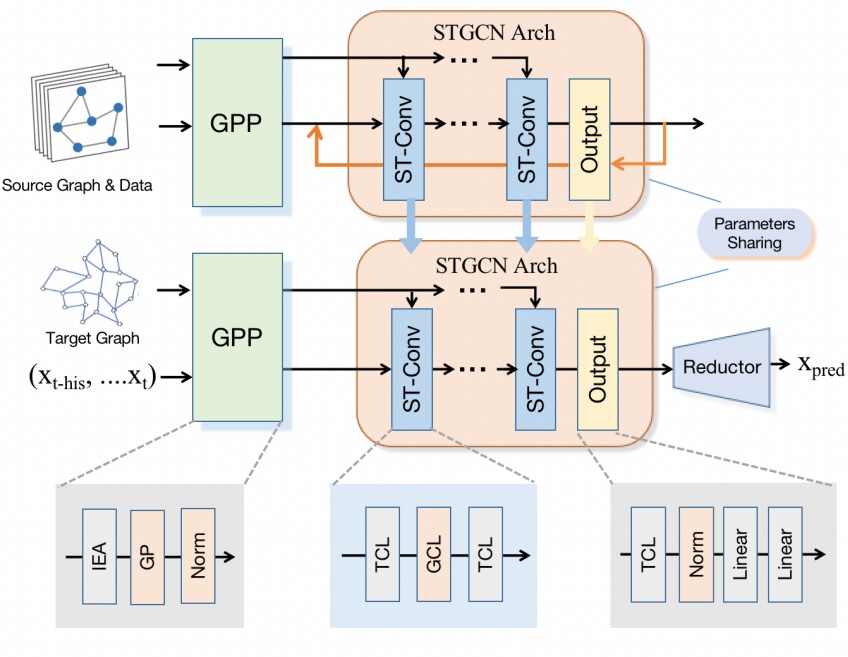}}
\caption{Model Structure of TL-GPSTGN. The framework utilizes a transfer learning approach with spatio-temporal graph convolutional networks (STGCN) for traffic prediction. The architecture processes source and target graph data through Graph Preprocessing Pipelines (GPP) and spatio-temporal convolutional layers (ST-Conv) with parameter sharing. The model outputs predictions ($X_{pred}$) using feature reductor layers, supported by input preprocessing (IEA, GP, Norm) and task-specific layers (TCL, GCL, Linear).}
\label{fig:structure}
\end{figure}

Firstly, source data with rich spatial-temporal features are processed by the GPP module, and the graph structure and traffic data suitable for transfer learning are extracted. Then, the pre-processed data are fed to the STGCN model for sufficient training and learning to obtain a pre-trained spatial-temporal graph convolutional model, which will share the learning parameters with the model used for the target road network. The graph structure and a small amount of training data in the target road network are likewise processed by GPP to become road network data with stronger characterization capabilities, which are fed into the spatial-temporal graph convolution model for the target road network for training to obtain the migrated pre- trained predictive model.

\subsection{STGCN Module}
Temporal Convolutional Layer (TCL) and Spatial Convolutional Layer (SCL) are two basic and important components of the STGCN model’s accepts the spatial-temporal data including the adjacency matrix of the graph and the feature matrix of the nodes and performs spatial convolutional operations on the input node feature data. The neighbor matrix is used to capture the spatial relationship between modestly accepting the feature matrix or the output of SCL to perform the convolution operation on it in the time dimension. The rational pairing of these two modules enables STGCN to combine the ability to process both spatial and temporal information and to effectively capture spatial-temporal relationships and features in graph networks \cite{b8}. Two TCLs and one SCL are combined in a "hamburger" fashion to form a spatial- temporal convolution module (ST-Conv), while the STGCN consists of several (usually two or three) ST-Conv modules and a data output processing module. The output processing module consists of a TCL to enhance the capturing of temporal features and is composed with several linear layers. The advantage of being able to simultaneously consider spatio-temporal information makes STGCN perform well in predicting complex road networks.

\subsection{Graph Pruning Processor}
Graph pruning is a technique used to reduce unnecessary connections and edges in a graph \cite{b9}. In graph pruning, a more simplified graph is obtained by removing certain edges and nodes from the graph while retaining the main feature structure and information of the graph. It can serve to remove redundant complexity and improve computational efficiency and migration capabilities. The common methods of graph pruning are Threshold-based Pruning, Degree-based Pruning, Centrality-based Pruning, Clustering-based Pruning, etc. Threshold Pruning is to remove edges whose weights are less than the threshold according to the given threshold value. Degree- based Pruning is to remove nodes with lower degrees or to remove weaker connections between nodes. Centrality-based Pruning is to remove nodes or edges whose centrality metrics are lower than the threshold. Clustering-based Pruning is to remove connections between clusters or with low density by clustering algorithms, which categorize the nodes in the graph into different clusters.

In traffic prediction, the road network graph has a strong coupling relationship with the feature data, and the nodes of the graph have a one-to-one mapping relationship with the dimensions of the feature data, so Degree-based Pruning is a more applicable method to select nodes and edges that have a greater contribution to the features of the graph as the input data to the model.

\begin{figure}[htbp]
\centerline{\includegraphics[width=\linewidth]{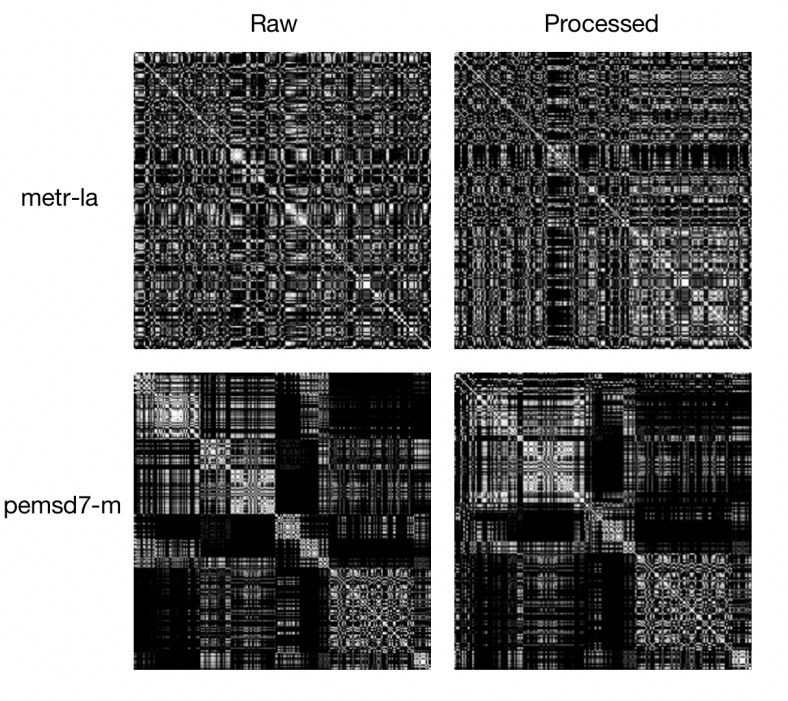}}
\caption{Graph Pruning Visualization. This figure shows the raw and processed adjacency matrices for the metr-la (top) and pemsd7-m (bottom) datasets. Graph pruning reduces noise and highlights significant connections, improving sparsity and enhancing the model's ability to capture meaningful spatial relationships.}
\label{fig:pruning}
\end{figure}

From Figure 2, it can be observed the change of graph neighbor matrix before and after pruning on metr-la and pemsd7-m datasets. In general, through the pruning process of the GPP module, the feature information of the graph will be more significant, and some nodes with less influence on the road network are filtered out. In the actual transportation road network, each road network region does not exist and operates independently, and different transportation regions are interconnected, so the road conditions of the studied road network are not only completely determined by the nodes and edges within that road network. In the edge region of the road network, there may be some nodes that are affected by the influence of the region outside the road network, and this influence is not taken into account by the prediction system, so the use of graph pruning technique can effectively reduce the participation of these nodes in the operation, which is very meaningful for improving the efficiency and performance of the model and the ability to migrate.

\subsection{Transfer Learning}
When graphical convolutional networks are applied in real transportation, a large amount of historical data is needed to support the pre-training of the model. Many road networks do not have enough historical data due to underdeveloped informatization and other reasons, thus creating the problem of rapid application of predictive models.

Transfer learning is a useful method affiliated to machine learning that focuses on storing solution models for existing scenes and exploiting them on other related problems. In the field of traffic prediction, the uniformity of the input data in roads allows transfer learning to play a significant role in it. Models can be trained on a road network with sufficient historical data and rich spatial features to obtain a pre-trained model. This model is then migrated to the target road network and trained a second time using less historical data to quickly obtain a model usable in the target domain \cite{b10} \cite{b11}.

\section{Experiment}
In this section, sufficient experiments are designed in this paper to demonstrate the superior performance and migration capability of the TL-GPSTGN model in traffic prediction tasks. First, the dataset, data preprocessing, and BASELINE method will be presented. Then, the model's performance and migration ability on different datasets will be demonstrated.

\subsection{Dataset and Preprocessing}
METR-LA (METRic for Traffic Flow Prediction in Los Angeles) provides highly accurate traffic flow data from the Los Angeles area. The dataset was provided by the Los Angeles Department of Transportation and the data was collected using Inductive Loop Detectors. The time frame of the dataset covers a one-year period from 2012 to 2013, with traffic flow data sampled at 10-minute intervals. It contains traffic flow data from 207 Inductive Loop Detectors.

The PEMS-BAY (PeMS Bay Area) dataset is a larger-scale traffic dataset for traffic flow covering the roadway network in the Bay Area (Bay Area) of California, USA. The dataset is created and maintained by the Intelligent Transportation Lab (ITS Lab) at the University of California, Berkeley. The dataset has been collected since 2007 and contains traffic flow data from over 300 traffic detectors. The traffic detectors are located on major roadways in the Bay Area, covering key locations such as freeways, city streets, and bridges, allowing the data to be used to monitor the overall condition of the roadway network. Each detector provides information on the flow of vehicles passing through at intervals of every minute.

PEMSD7 is a large-scale real-time traffic dataset used in the field of traffic prediction, created by the Intelligent Transportation Laboratory (ITS Lab) at the University of California, Berkeley. The dataset is based on PeMS (Performance Measurement System) traffic data from the state of California and contains real-time traffic flow data from 200 traffic detectors, providing traffic flow information at five- minute intervals.

To transform the data into a standard normal distribution with a specific mean and standard deviation, standardization, a data preprocessing technique, is employed. It scales each feature to zero mean and unit variance by applying a linear transformation to it. This eliminates the differences in magnitude between features and helps to improve data comparability and model stability.

It is worth noting that for the traffic prediction dataset, the normalization operation during preprocessing is performed based on the statistics of the training dataset, and then the same transformations are applied to the validation and test datasets to ensure that the prediction and evaluation are performed with the same data distribution. That is, the mean and standard deviation of the training set are used to standardize the validation and test sets to maintain consistency and reproducibility of data processing.

\subsection{Evaluation Metrics and Baseline}
In the model evaluation in this paper, Mean Absolute Error (MAE), Root Mean Square Error (RMSE), and Mean Absolute Percentage Error (MAPE) are used as the metrics to evaluate the performance of TL-GPSTGN \cite{b11}.

HA (Historical Average): A forecast based on the average of historical observations. It assumes that future values are similar to past observations and takes the historical average as the forecast.

ARIMA (Autoregressive Integrated Moving Average): the classical time series forecasting model. It combines the concepts of autoregressive (AR) and moving average (MA) models and takes into account the differencing (Integrated) process of the time series \cite{b12}.

FNN (Feedforward Neural Network): a classical artificial neural network model. Models complex nonlinear relationships between inputs and outputs by learning and adjusting connection weights between neurons \cite{b13}.

FC-LSTM (Fully-Connected LSTM): a neural network model based on Long Short-Term Memory (LSTM). It captures the long-term dependencies of the time series through LSTM layers and uses a fully-connected layer in the last layer for prediction \cite{b14}.

By using these models as evaluation benchmarks and using MAE, RMSE, and MAPE as evaluation metrics, it can be well evaluated whether the TL-GPSTGN model has good prediction accuracy and migration performance in the field of traffic prediction.

\subsection{Experiments and Results}
To ensure the uniform standard of the evaluation, all the evaluation tasks are conducted on two machines equipped with RTX4090 graphics cards and 24G video memory, and the whole experiment takes about 72 hours in total. The evaluation tasks are mainly divided into two modules, one is the performance evaluation of TF- GPSTGN model on a single dataset, and the other is the performance evaluation of TL-GPSTGN migration between different datasets. In the single dataset evaluation, three datasets, metr-la, pems-bay, and pemsd7-m, are selected in this paper, and comparative experiments are done with HA, ARIMA, FNN, FC-LSTM, and STGCN under the same conditions to validate the accuracy of the TL-GPSTGN model. In the migration performance evaluation, the STGCN model, which has the best performance in the single dataset evaluation, is selected to do the comparison experiment with TL-GPSTGN. In this task, the two models are pre-trained sufficiently on the metr-la dataset respectively, and then a part of the pems-bay and pemsd7-m datasets are selected respectively for further migration training, and the performance evaluation is conducted on the migrated dataset to get the migration performance of the models.

Training parameter settings. The three datasets are divided into the training set, validation set, and test set in the ratio of 7:1.5:1.5 respectively. Early Stopping technique is used, and the maximum value of epoch is set to 200. The length of the predicted historical data sequence is 12 and the prediction intervals are 15 and 30 minutes. Since the pems-bay dataset has a shorter sampling interval of 1 minute, specifically, the input sequence length for the 30-minute prediction task for this dataset is set to 24. The random seed is set to 42, the number of ST-Conv modules in the STGCN structure is set to 2, and the convolution kernel sizes of the SCL and TCL convolution modules in it are set to 3. The batch size is set to 32, the learning rate is set to 0.001, the L2 regularization penalty is used, the weight decay ratio is set to 0.0005, and the optimizer uses Adam.

In the single dataset evaluation, the predictions of STGCN and TL-GPSTGN are tested on three datasets against 15 and 30 minutes, respectively, and the corresponding performance metrics are calculated. Then they are compared with the baseline models and the results are shown in the table.

\begin{table}[htbp]
\centering
\caption{Results of single dataset evaluation (15 minutes)}
\label{tab:single_dataset_results_15}
\renewcommand{\arraystretch}{1.0}
\begin{adjustbox}{max width=0.49\textwidth}
\begin{tabular}{@{}lccccccccc@{}}
\toprule
\textbf{Models} & \multicolumn{3}{c}{\textbf{metr-la}} & \multicolumn{3}{c}{\textbf{pems-bay}} & \multicolumn{3}{c}{\textbf{pemsd7}} \\ 
\cmidrule(lr){2-4} \cmidrule(lr){5-7} \cmidrule(lr){8-10}
 & \textbf{MAE} & \textbf{MAPE} & \textbf{RMSE} & \textbf{MAE} & \textbf{MAPE} & \textbf{RMSE} & \textbf{MAE} & \textbf{MAPE} & \textbf{RMSE} \\ 
\midrule
HA & 4.16 & 13.0 & 7.80 & 2.88 & 6.80 & 5.59 & 4.01 & 10.61 & 4.55 \\ 
ARIMA & 3.99 & 9.6 & 8.21 & 1.62 & 3.50 & 3.30 & 5.55 & 12.92 & 9.00 \\ 
FNN & 3.99 & 9.9 & 7.94 & 2.20 & 5.19 & 4.42 & 2.74 & 6.38 & 4.75 \\ 
FC-LSTM & 3.44 & 9.6 & 6.30 & 2.05 & 4.8 & 4.19 & 3.57 & 8.60 & 6.20 \\ 
STGCN & 3.40 & 6.71 & 6.56 & 2.50 & 4.03 & 5.11 & 2.38 & 4.16 & 4.29 \\ 
TL-GPSTGN & 3.37 & 6.58 & 6.62 & 2.92 & 4.77 & 5.79 & 2.47 & 4.34 & 4.30 \\ 
\bottomrule
\end{tabular}
\end{adjustbox}
\end{table}

\begin{table}[htbp]
\centering
\caption{Results of single dataset evaluation (30 minutes)}
\label{tab:single_dataset_results_30}
\renewcommand{\arraystretch}{1.0}
\begin{adjustbox}{max width=0.49\textwidth}
\begin{tabular}{@{}lccccccccc@{}}
\toprule
\textbf{Models} & \multicolumn{3}{c}{\textbf{metr-la}} & \multicolumn{3}{c}{\textbf{pems-bay}} & \multicolumn{3}{c}{\textbf{pemsd7}} \\ 
\cmidrule(lr){2-4} \cmidrule(lr){5-7} \cmidrule(lr){8-10}
 & \textbf{MAE} & \textbf{MAPE} & \textbf{RMSE} & \textbf{MAE} & \textbf{MAPE} & \textbf{RMSE} & \textbf{MAE} & \textbf{MAPE} & \textbf{RMSE} \\ 
\midrule
HA & 4.16 & 13.0 & 7.80 & 2.88 & 6.80 & 5.59 & 4.01 & 10.61 & 6.67 \\ 
ARIMA & 5.15 & 12.7 & 10.45 & 2.33 & 8.30 & 4.76 & 5.86 & 13.94 & 9.13 \\ 
FNN & 4.23 & 12.9 & 8.17 & 2.30 & 5.43 & 4.63 & 4.02 & 9.72 & 6.98 \\ 
FC-LSTM & 3.77 & 10.9 & 7.23 & 2.20 & 5.2 & 4.55 & 3.94 & 9.55 & 7.03 \\ 
STGCN & 4.41 & 8.71 & 9.17 & 3.26 & 5.26 & 6.59 & 3.05 & 5.27 & 5.22 \\ 
TL-GPSTGN & 4.45 & 8.70 & 9.34 & 3.51 & 5.76 & 6.67 & 3.18 & 5.53 & 5.72 \\ 
\bottomrule
\end{tabular}
\end{adjustbox}
\end{table}

\begin{figure}[htbp]

\centering
\begin{subfigure}{0.24\textwidth}
    \includegraphics[width=\linewidth]{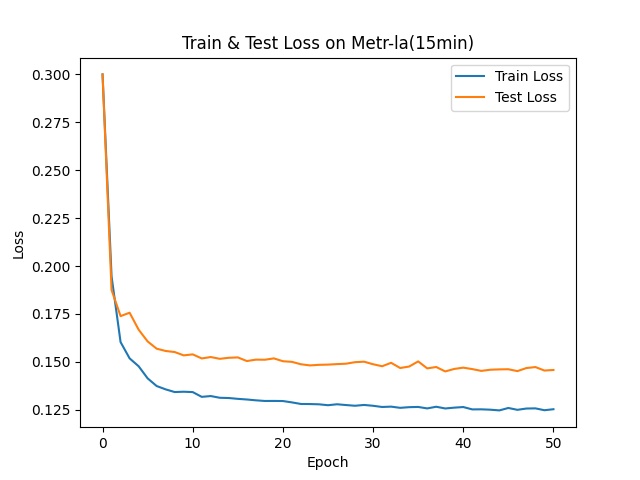}
    \label{fig1}
\end{subfigure}
\hfill
\begin{subfigure}{0.24\textwidth}
    \includegraphics[width=\linewidth]{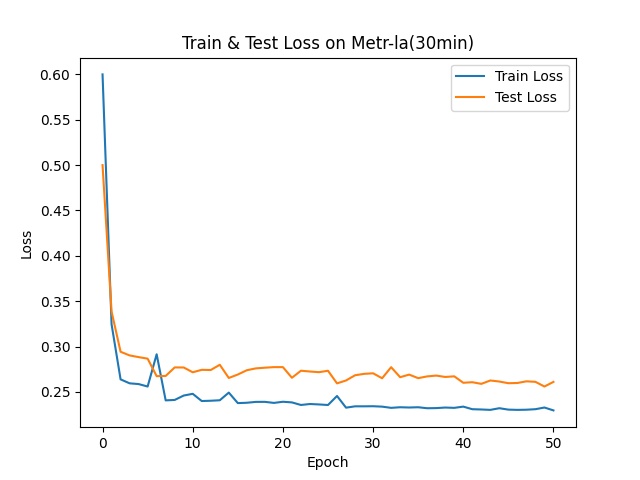}
    \label{fig2}
\end{subfigure}
\small \caption{Loss of TL-GPSTGN trained on metr-la dataset. Train and test loss for 15-minute (left) and 30-minute (right) prediction intervals. Both plots show rapid loss reduction in early epochs, followed by stabilization, highlighting the model's convergence and generalization ability.}
\label{fig}
\end{figure}

As can be seen from Table~\ref{tab:single_dataset_results_15}, Table~\ref{tab:single_dataset_results_30} and Figure 3, both the STGCN model and the TL- GPSTGN model significantly outperform the other four baseline models in all three datasets. The TL-GPSTGN model slightly outperforms the STGCN model in terms of prediction accuracy on the metr-la dataset and weakly outperforms it on the other two datasets. In conclusion, the prediction performance of the TL-GPSTGN model on a single dataset is comparable to that of the STGCN model and significantly better than the other models. This indicates that TL-GPSTGN is effective in extracting the main features of the graph after GPP processing of the source feature data and graph structure, and can achieve and maintain good prediction performance even when the graph structure complexity and feature data are reduced.

Model migration performance test. In this test task, in this paper, the STGCN and TL-GPSTGN models are fully trained on the metr-la dataset before migrating to the pemsd7-m and pems-bay datasets for migration training and testing, respectively. For the pemsd7-m dataset, the two models were tested for 15-minute and 30-minute prediction performance, respectively, and 15-minute prediction test on the other dataset. The Target/Source parameter, which is the data ratio of the target dataset and the source dataset used, was used to evaluate the migration prediction performance of the models with different ratios of old and new data.

\begin{figure}[htbp]
\centering
\begin{subfigure}{0.24\textwidth}
    \includegraphics[width=\linewidth]{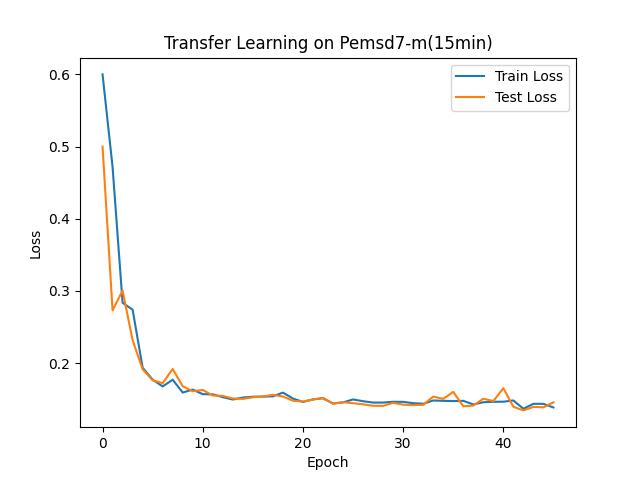}
    \label{fig1}
\end{subfigure}
\hfill
\begin{subfigure}{0.24\textwidth}
    \includegraphics[width=\linewidth]{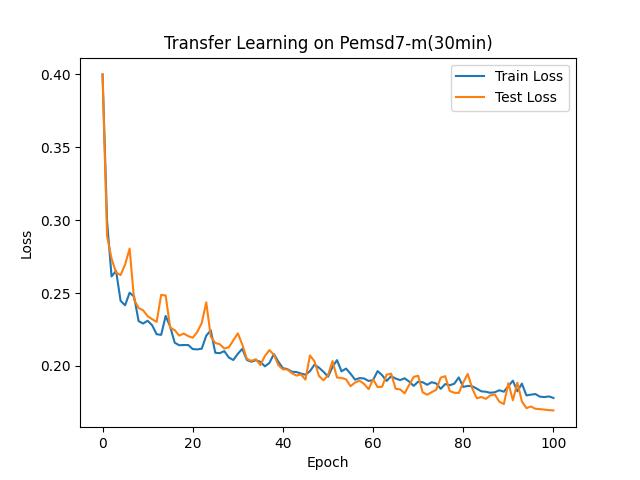}
    \label{fig2}
\end{subfigure}
\caption{Transfer learning loss on pemsd7-m dataset. Loss curves for transfer learning on 15-minute (left) and 30-minute (right) intervals. Both plots demonstrate the model's effective adaptation to the target dataset, with steady loss reduction and convergence, showcasing successful knowledge transfer.}
\label{fig}
\end{figure}

\begin{table}[htbp]
\centering
\caption{Transfer capability evaluation of different models}
\label{tab:transfer_capability}
\renewcommand{\arraystretch}{1.0}
\begin{adjustbox}{max width=0.5\textwidth}
\begin{tabular}{@{}llccccccccc@{}}
\toprule
\textbf{Model} & \textbf{T/S} & \multicolumn{3}{c}{\textbf{Metr-la To pemsd7-m}} & \multicolumn{3}{c}{\textbf{Metr-la To pems-bay}} \\ 
\cmidrule(lr){3-5} \cmidrule(lr){6-8}
 & & \textbf{MAE} & \textbf{MAPE (\%)} & \textbf{RMSE} & \textbf{MAE} & \textbf{MAPE (\%)} & \textbf{RMSE} \\ 
 & & (15/30min) & (15/30min) & (15/30min) & (15min) & (15min) & (15min) \\ 
\midrule
STGCN & 5\%  & 3.34/4.80 & 5.83/8.36 & 5.65/8.37 & 4.71 & 7.73 & 9.40 \\ 
      & 10\% & 3.24/4.48 & 5.65/7.79 & 5.27/7.50 & 4.17 & 6.84 & 7.54 \\ 
      & 15\% & 2.83/4.00 & 4.93/6.96 & 4.76/6.73 & 4.29 & 7.04 & 7.18 \\ 
      & 20\% & 2.76/3.92 & 4.81/6.82 & 4.74/6.70 & 4.11 & 6.74 & 7.11 \\ 
      & 25\% & 2.74/3.63 & 4.77/6.41 & 4.67/6.47 & 4.69 & 7.70 & 7.64 \\ 
GPSTG & 5\%  & 3.22/4.40 & 5.52/7.55 & 5.51/7.79 & 3.57 & 5.76 & 7.41 \\ 
      & 10\% & 3.09/4.37 & 5.29/7.51 & 5.17/7.40 & 3.58 & 5.76 & 7.20 \\ 
      & 15\% & 2.56/3.50 & 4.39/6.01 & 4.44/6.12 & 3.90 & 6.27 & 6.79 \\ 
      & 20\% & 2.49/3.67 & 4.28/6.29 & 4.43/6.29 & 3.10 & 5.00 & 5.79 \\ 
      & 25\% & 2.56/3.41 & 4.38/5.84 & 4.43/6.07 & 3.07 & 4.94 & 5.80 \\ 
\bottomrule
\end{tabular}
\end{adjustbox}
\end{table}

As can be seen from Table~\ref{tab:transfer_capability} and Figure 4, as the ratio of target data to source data increases, the performance of the two models on the new dataset becomes progressively better, even approaching the performance on the single target dataset. This indicates that both models have more stable performance in the migration task. Meanwhile, the TL-GPSTGN model comprehensively outperforms the STGCN model both in different datasets and in different prediction intervals of the same dataset. This indicates that the TL-GPSTGN model is indeed able to extract the main features in the road network that affect the prediction results, improve the prediction efficiency and accuracy, and have excellent migration performance.

Migration prediction results are shown. to be able to further verify the prediction performance of the TL-GPSTGN model and the value of its application to the actual road network, this paper outputs the model's prediction data to be processed into the actual traffic prediction results and graphically demonstrates them at 10-minute intervals.

\begin{figure}[htbp]
\centering
\begin{subfigure}{0.24\textwidth}
    \includegraphics[width=\linewidth]{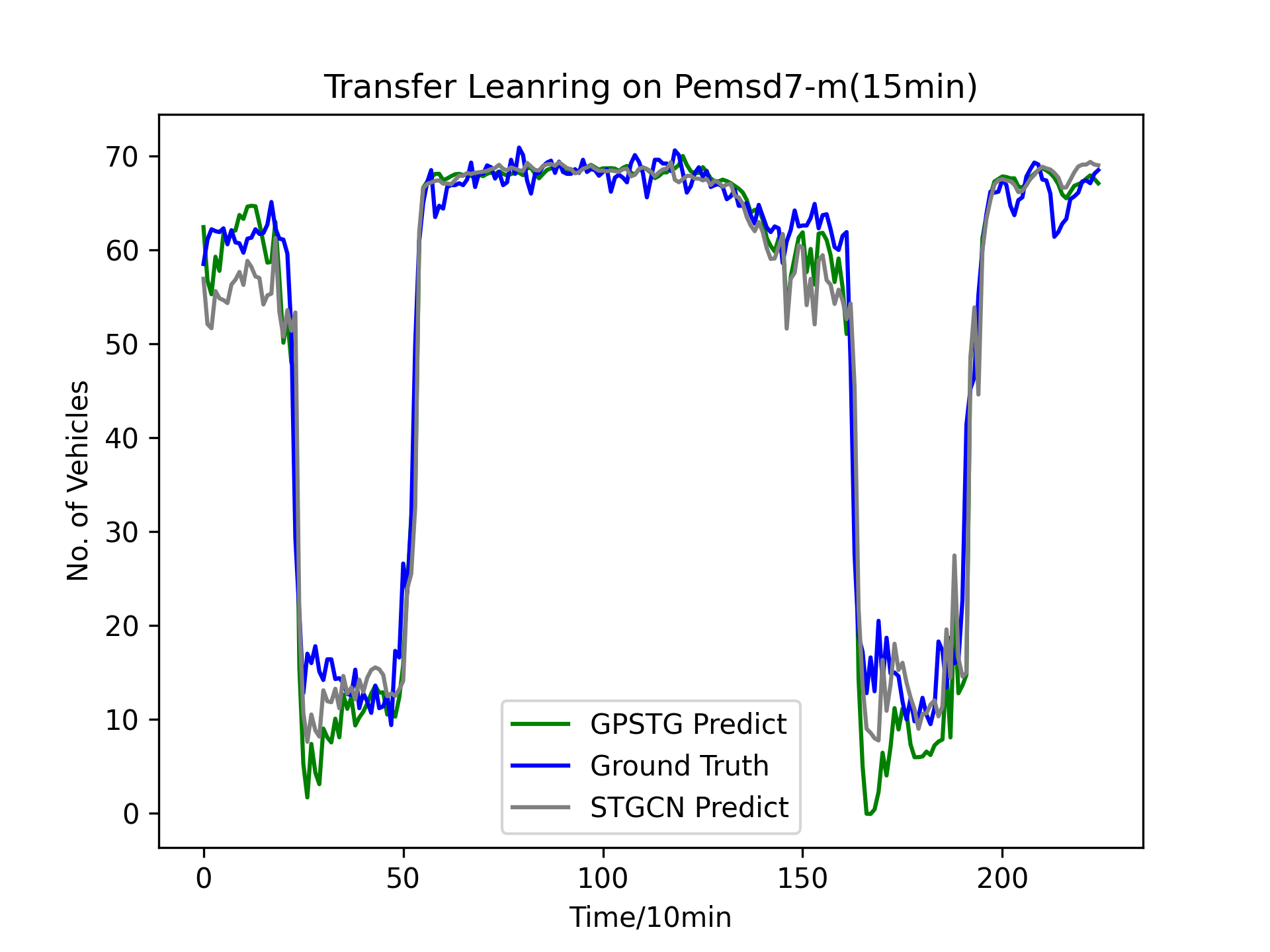}
    \label{fig1}
\end{subfigure}
\hfill
\begin{subfigure}{0.24\textwidth}
    \includegraphics[width=\linewidth]{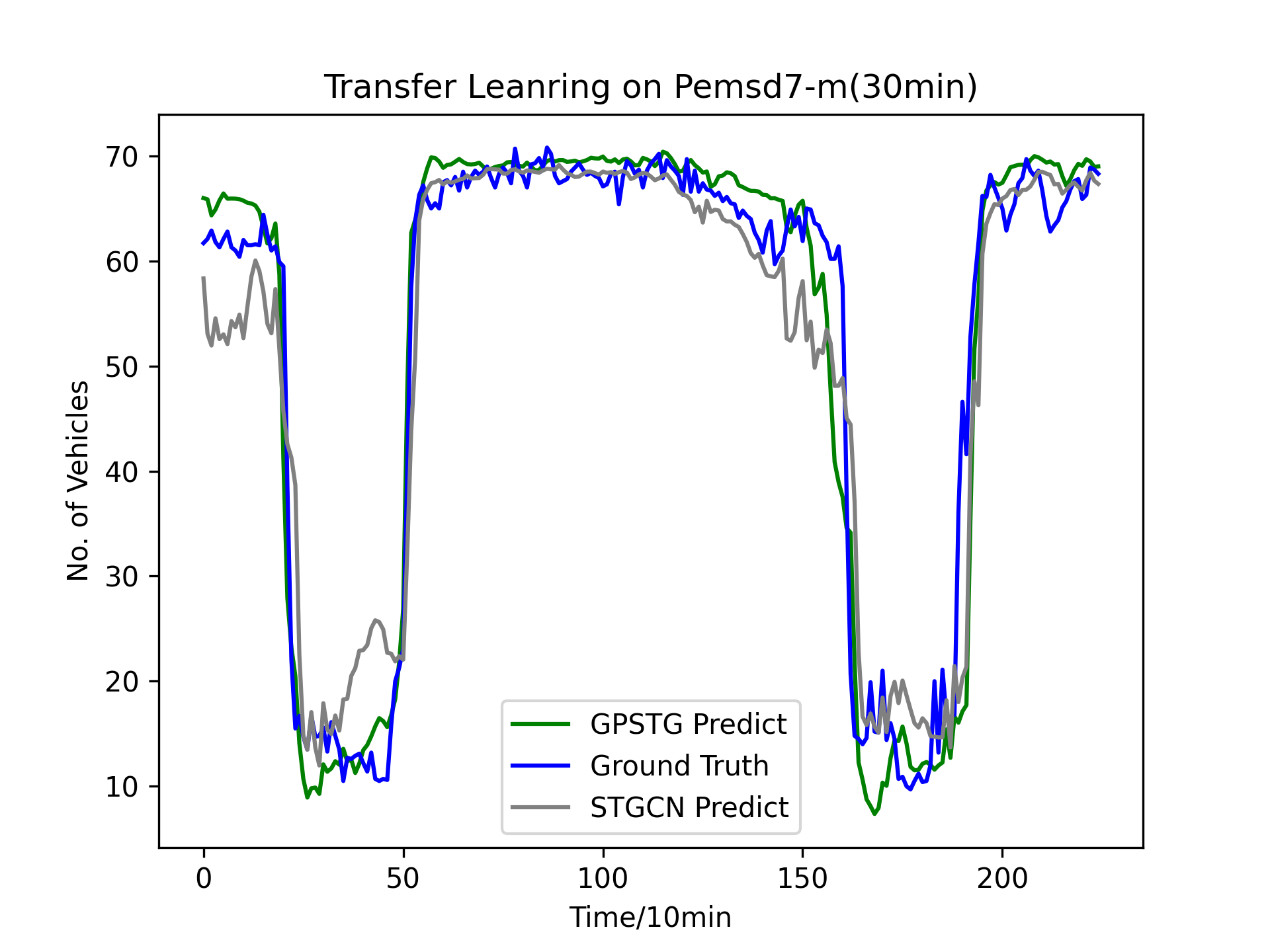}
    \label{fig2}
\end{subfigure}
\caption{Partial prediction results on the pemsd7-m datasets. Predictions for 15-minute (left) and 30-minute (right) intervals on the Pemsd7-m dataset using transfer learning. The GPSTG model aligns closely with the ground truth, outperforming the baseline STGCN model in capturing traffic patterns.}
\label{fig}
\end{figure}

\begin{figure}[htbp]
\centering
\begin{subfigure}{0.24\textwidth}
    \includegraphics[width=\linewidth]{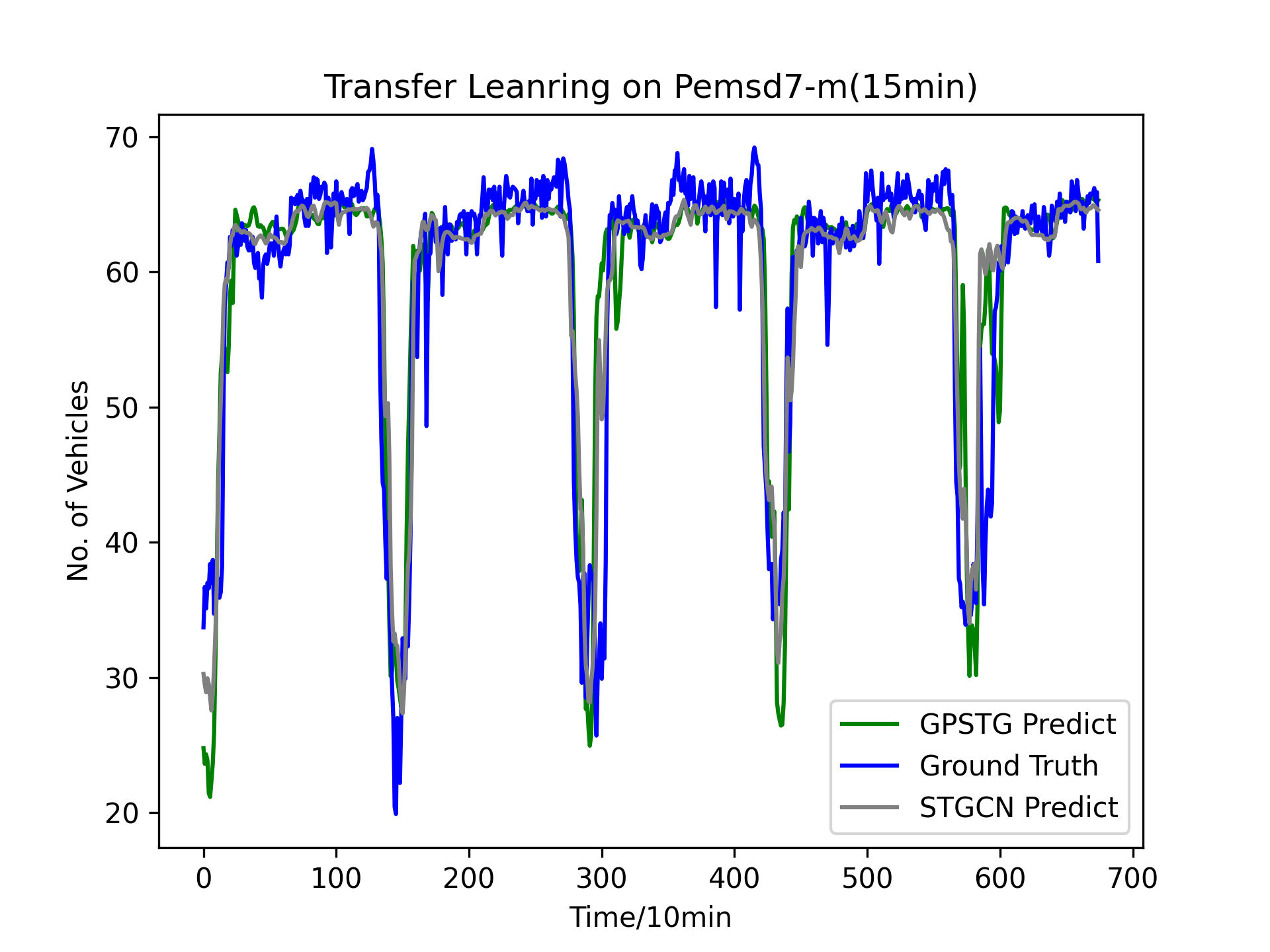}
    \label{fig1}
\end{subfigure}
\hfill
\begin{subfigure}{0.24\textwidth}
    \includegraphics[width=\linewidth]{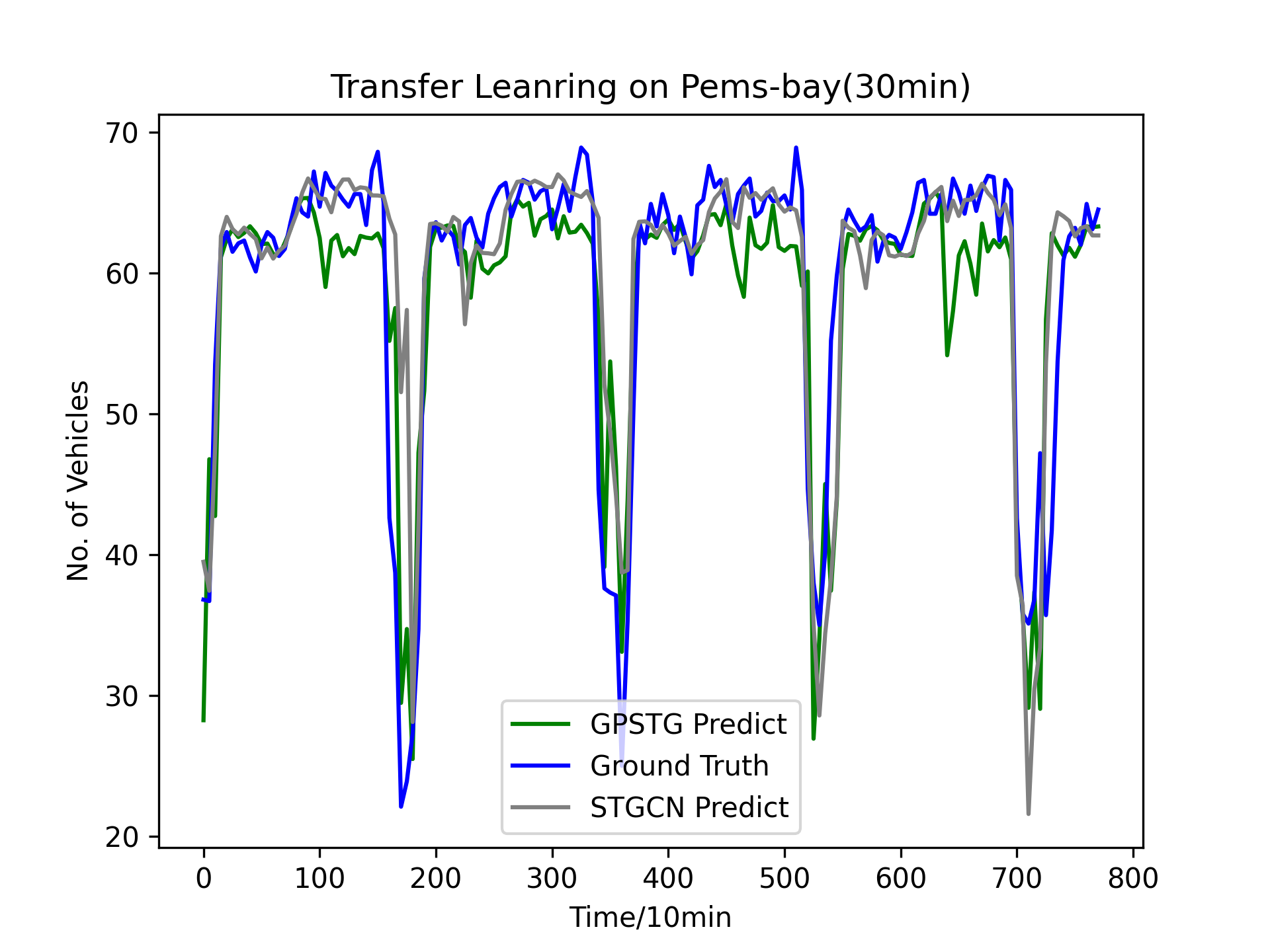}
    \label{fig2}
\end{subfigure}
\caption{Partial prediction results on the pems-bay datasets. Predictions for 15-minute (left) and 30-minute (right) intervals on the Pems-bay dataset using transfer learning. The GPSTG model demonstrates strong predictive performance and adapts well to the dataset, closely following the ground truth compared to STGCN.}
\label{fig}
\end{figure}

As can be seen from Figure 5 and Figure 6, both models fit better for 15-minute predictions. Among them, the TL-GPSTGN model is able to predict more accurately for the important extremely large and extremely small points and also has more accurate fitting results for the dramatic data rise and fall. Therefore, TL-GPSTGN can predict the changes of traffic flow well in real traffic prediction, and it can also predict the values of traffic peaks and troughs well.

The experiments conducted in this study demonstrate the superior prediction performance and migration capability of the TL-GPSTGN model in traffic prediction tasks. Through evaluations on three real-world datasets (metr-la, pems-bay, and pemsd7-m), the TL-GPSTGN model consistently outperformed baseline models, including HA, ARIMA, FNN, FC-LSTM, and STGCN, in both single-dataset and transfer learning scenarios. The migration experiments revealed that TL-GPSTGN effectively adapts to new road networks with minimal target data while maintaining high accuracy. Visualization of prediction results further highlights its ability to capture critical traffic patterns, including peaks and troughs, making it a robust tool for real-world traffic forecasting and smart transportation systems. These findings validate the model's efficiency, scalability, and applicability across diverse traffic environments.

\section{Generalization Analysis}
In this section, we delve into the Rademacher Complexity analysis to evaluate the model's generalization capabilities. Rademacher Complexity serves as a theoretical framework to quantify the capacity of a hypothesis class to fit random noise, thereby providing insights into the model's ability to generalize to unseen data. Specifically, we focus on computing the Rademacher Complexity for convolution layers, temporal convolution layers, and the entire network, breaking down the contributions of various components, including align layers, causal convolutions, and activation functions. Additionally, we analyze how parameter sharing and weight regularization techniques impact complexity, revealing trade-offs between model expressiveness and overfitting. This analysis offers a comprehensive understanding of how the architectural components influence the overall complexity and generalization behavior of the network, highlighting potential avenues for optimization in future designs.

\subsection{Rademacher Complexity for Convolution layer}

\subsubsection{Recall the Definition of Rademacher Complexity}
The Rademacher Complexity of a hypothesis class $\mathcal{H}$ with respect to a sample set $S = \{x_1, x_2, \dots, x_m\}$ is defined as:
\[
\mathcal{R}_S(\mathcal{H}) = \mathbb{E}_{\sigma} \left[ \sup_{h \in \mathcal{H}} \frac{1}{m} \sum_{i=1}^m \sigma_i h(x_i) \right],
\]
where:
\begin{itemize}
    \item $\sigma = (\sigma_1, \sigma_2, \dots, \sigma_m)$ are independent and identically distributed Rademacher random variables, $\sigma_i \sim \{-1, 1\}$ with $\mathbb{P}(\sigma_i = 1) = \mathbb{P}(\sigma_i = -1) = 0.5$.
    \item $h(x_i)$ represents the output of the function $h$ applied to the input $x_i$.
\end{itemize}

For a convolution layer, the hypothesis $h(x)$ is parameterized by the convolution weights $W_{\text{conv}}$, and the Rademacher Complexity quantifies the capacity of the model to fit random noise.

\subsubsection{The Convolution Layer}
The output of the convolution layer can be expressed as:
\[
h(x) = \langle W_{\text{conv}}, \Phi(x) \rangle,
\]
where: $W_{\text{conv}} \in \mathbb{R}^{C_{\text{out}} \times C_{\text{in}} \times K_h \times K_w}$ is the weight matrix of the convolution layer. $\Phi(x)$ represents the local receptive field of the input $x$ associated with the convolution operation.

\subsubsection{Rademacher Complexity for the Convolution Layer}
Substituting the convolutional hypothesis into the Rademacher Complexity definition:
\begin{align*}
\mathcal{R}_S(\text{Conv}) &= \mathbb{E}_{\sigma} \left[ \sup_{W_{\text{conv}}} \frac{1}{m} \sum_{i=1}^m \sigma_i \langle W_{\text{conv}}, \Phi(x_i) \rangle \right] \\
    &= \frac{1}{m} \mathbb{E}_{\sigma} \left[ \sup_{W_{\text{conv}}} \langle W_{\text{conv}}, \sum_{i=1}^m \sigma_i \Phi(x_i) \rangle \right],
\end{align*}

Using the Cauchy-Schwarz inequality for the inner product:
\[
\langle W_{\text{conv}}, \sum_{i=1}^m \sigma_i \Phi(x_i) \rangle \leq \|W_{\text{conv}}\|_F \cdot \left\| \sum_{i=1}^m \sigma_i \Phi(x_i) \right\|_F,
\]
the $\| \cdot \|_F$ is the Frobenius norm. Substituting this into the Rademacher Complexity formula, we can get:
\[
\mathcal{R}_S(\text{Conv}) \leq \frac{\|W_{\text{conv}}\|_F}{m} \mathbb{E}_{\sigma} \left[ \left\| \sum_{i=1}^m \sigma_i \Phi(x_i) \right\|_F \right].
\]

We can also simplify the Expectation over Rademacher Variables.
The random variables $\sigma_i$ are independent and uniformly distributed, and for normalized inputs (e.g., $\|\Phi(x_i)\|_F \leq 1$), the expectation can be bounded as:
\[
\mathbb{E}_{\sigma} \left[ \left\| \sum_{i=1}^m \sigma_i \Phi(x_i) \right\|_F \right] \leq \sqrt{m}.
\]

As a result, we get the upper bound complexity of the convolution layer.
\[
\mathcal{R}_S(\text{Conv}) \leq \frac{\|W_{\text{conv}}\|_F}{m} \cdot \sqrt{m} = \frac{\|W_{\text{conv}}\|_F}{\sqrt{m}}.
\]

\subsection{Rademacher Complexity Calculation for TemporalConvLayer}

Temporal convolutional layer plays a crucial role in capturing temporal dependencies across time-series data in our model.

\subsubsection*{Basic Settings}

\begin{itemize}
    \item Final activation function using ReLU, with Lipschitz constant $L_{\text{ReLU}} = 1$.
    \item Align layer is a 2D convolution with kernel size $1 \times 1$.
    \item Input tensor should be in shape: $x \in \mathbb{R}^{m \times C_{\text{in}} \times T \times N}$.
    \item Weights for the Align layer: $W_{\text{align}} \in \mathbb{R}^{C_{\text{out}} \times C_{\text{in}} \times 1 \times 1}$.
    \item Weights for the causal convolution: $W_{\text{conv}} \in \mathbb{R}^{2 \times C_{\text{out}} \times C_{\text{in}} \times K_t \times 1}$.
    \item Activation functions are Lipschitz continuous.
    \item Rademacher complexity of the input tensor $x$ is $\mathcal{R}(x)$.
\end{itemize}

With the basic clarification of each component, we can calculate the Rademacher Complexity step by step:

\paragraph{Rademacher Complexity of Align Layer}
The Align layer is a $1 \times 1$ convolution, which can be treated as a linear transformation. 
The Rademacher Complexity of a linear transformation with weight matrix $W_{\text{align}}$ is bounded by:
\[
\mathcal{R}(\text{Align}) \leq \frac{\| W_{\text{align}} \|_F}{\sqrt{m}},
\]
where $\| W_{\text{align}} \|_F$ is the Frobenius norm of the weight matrix.

\paragraph{Rademacher Complexity of Causal Convolution Layer}
The causal convolution layer applies a kernel of size $K_t \times 1$ over the input. The output complexity depends on the Frobenius norm of the convolution weights $W_{\text{conv}}$:
\[
\mathcal{R}(\text{CausalConv}) \leq \frac{\| W_{\text{conv}} \|_F}{\sqrt{m}}.
\]

\paragraph{Contribution of the Activation Function (ReLU)}
The ReLU activation function is 1-Lipschitz, meaning it does not increase the Rademacher Complexity. Thus, the complexity remains:
\[
\mathcal{R}(\text{ReLU}) = \mathcal{R}(\text{CausalConv}).
\]

\paragraph{Combining Align Layer, Causal Convolution, and ReLU}
The TemporalConvLayer consists of a residual connection that combines the output of the Align layer and the causal convolution, followed by the ReLU activation. 
The Rademacher Complexity of the layer is the sum of the contributions:
\[
\mathcal{R}(\text{TemporalConvLayer}) \leq \mathcal{R}(\text{Align}) + \mathcal{R}(\text{CausalConv}).
\]
Substituting the bounds for the individual layers:
\[
\mathcal{R}(\text{TemporalConvLayer}) \leq \frac{\| W_{\text{align}} \|_F}{\sqrt{m}} + \frac{\| W_{\text{conv}} \|_F}{\sqrt{m}}.
\]

The final Rademacher Complexity of the TemporalConvLayer is given by:
\[
\mathcal{R}(\text{TemporalConvLayer}) \leq \frac{1}{\sqrt{m}} \left( \| W_{\text{align}} \|_F + \| W_{\text{conv}} \|_F \right).
\]

Now we can set some parameters to estimate  \( \mathcal{R}(\text{TemporalConvLayer}) \).

In the experiments, we set the temporal kernel size: $K_t = 3$. The input channels $C_{\text{in}} = 16$, with the output channels $C_{\text{out}} = 32$; Sample size: $m = 10000$; The number of vertices $N = 207$. 

The Frobenius norm of a \(1 \times 1\) convolution kernel \( W \in \mathbb{R}^{C_{\text{out}} \times C_{\text{in}} \times 1 \times 1} \) is defined as:
\[
\| W \|_F = \sqrt{\sum_{i=1}^{C_{\text{out}}} \sum_{j=1}^{C_{\text{in}}} W_{i,j}^2}.
\]

For the weights of the Align layer, we use Xavier initialization, the weights are sampled from a uniform distribution:
\[
W_{i,j} \sim U\left(-\sqrt{\frac{6}{C_{\text{in}} + C_{\text{out}}}}, \sqrt{\frac{6}{C_{\text{in}} + C_{\text{out}}}}\right).
\]
The variance of each weight is:
\[
\text{Var}(W_{i,j}) = \frac{2}{C_{\text{in}} + C_{\text{out}}}.
\]

The expected squared Frobenius norm is given by:

\[
\mathbb{E}[\| W \|_F^2] = C_{\text{out}} \cdot C_{\text{in}} \cdot \mathbb{E}[W_{i,j}^2],
\]
where:
\[
\mathbb{E}[W_{i,j}^2] = \text{Var}(W_{i,j}) = \frac{2}{C_{\text{in}} + C_{\text{out}}}.
\]

Thus:
\[
\mathbb{E}[\| W \|_F^2] = C_{\text{out}} \cdot C_{\text{in}} \cdot \frac{2}{C_{\text{in}} + C_{\text{out}}}.
\]

Taking the square root, the expected Frobenius norm is:
\[
\mathbb{E}[\| W \|_F] = \sqrt{C_{\text{out}} \cdot C_{\text{in}} \cdot \frac{2}{C_{\text{in}} + C_{\text{out}}}}.
\]

Let's bring in the specific numbers, for \( C_{\text{in}} = 16 \) and \( C_{\text{out}} = 32 \), In the end, we get the expected Frobenius norm of the Align layer is approximate:

\begin{align*}
    \mathbb{E}[\| W_{align} \|_F] &= \sqrt{32 \cdot 16 \cdot \frac{2}{16 + 32}} = \sqrt{21.333} \approx 4.62  \\
\end{align*}

Using the same procedure, we can calculate the Frobisher norm of the casual convolution layer:

\[
\mathbb{E}[\|W_{Conv}\|_F] = \sqrt{32 \cdot 16 \cdot 3 \cdot 1 \cdot \frac{2}{16 + 32}} \approx 8.01.
\]

The Temporal Convolution layer consists of:
\begin{align*}
    \mathcal{R}(\text{TemporalConvLayer}) &= \mathcal{R}(\text{Align}) + \mathcal{R}(\text{CausalConv}),
\end{align*}
where:
\begin{align*}
    \mathcal{R}(\text{Align}) &\leq \frac{\| W_{\text{align}} \|_F}{\sqrt{m}} = \frac{4.62}{\sqrt{10000}} =  0.0462 \\
    \mathcal{R}(\text{CausalConv}) &\leq \frac{\| W_{\text{Conv}} \|_F}{\sqrt{m}} = \frac{8.01}{\sqrt{10000}} = 0.0801
\end{align*}

\begin{align*}
    \mathcal{R}(\text{TemporalConvLayer}) &\leq 0.0462 + 0.0801 \approx 0.13.
\end{align*}

\subsection{Rademacher Complexity Calculation for the Network}

For the backbone of the model, we set the number of main blocks to 2. Each block has 3 convolutional layers similar to the structure of the temporal layer.  

For a network consisting of $L$ identical blocks, each with Rademacher Complexity $\mathcal{R}(\text{Block})$, and Lipschitz constant $L_{\text{Block}}$, the Rademacher Complexity of the entire network is bounded by:
\[
\mathcal{R}(\text{Network}) \leq \prod_{l=1}^L L_{\text{Block}} \cdot \mathcal{R}(\text{Block}),
\]

The total Lipschitz constant of a block with two convolution layers and activation functions is:
\[
L_{\text{Block}} = L_{\text{align}} \cdot L_{\text{act1}} \cdot L_{\text{Conv}}
\]

If the activation functions are ReLU, their Lipschitz constants are $L_{\text{act1}} = L_{\text{act2}} = 1$. Therefore:
\[
L_{\text{Block}} = L_{\text{conv1}} \cdot L_{\text{conv2}}.
\]

For convolution layers, the Lipschitz constants are approximated using the Frobenius norm of the weights:
\[
L_{\text{conv1}} = \|W_1\|_F = 4.62, \quad L_{\text{conv2}} = \|W_2\|_F = 8.01.
\]

Substituting these into the equation:
\[
L_{\text{Block}} = \|W_1\|_F \cdot \|W_2\|_F = 37.01.
\]

So the Rademacher Complexity of the network's backbone will be:

\[
\mathcal{R}(\text{Network}) \leq 37.01 \cdot 0.13 = 4.81
\]

Through the detailed computation and analysis, we have established an upper bound for the Rademacher Complexity of the network, reflecting its generalization potential. By systematically considering the contributions of convolution layers, align layers, and activation functions, we demonstrated how the architecture's design impacts its capacity to fit data while maintaining robustness. The derived results underline the importance of architectural choices in balancing model expressiveness with generalization, offering valuable guidance for designing efficient and scalable models with strong generalization performance.

\section{Conclusion}
In this paper, we introduced TL-GPSTGN, a spatial-temporal graph convolutional network model leveraging transfer learning and graph pruning to tackle the challenge of limited historical data in traffic prediction. By transferring knowledge from source road networks with rich data to target networks with sparse data, TL-GPSTGN enhances prediction accuracy. The model integrates data source relevance and information entropy analyses, employs graph pruning to eliminate low-relevance nodes and connections, and utilizes an STGCN module to extract spatial-temporal features for training and prediction.

Extensive evaluations on three real-world datasets demonstrate that TL-GPSTGN outperforms traditional models (HA, ARIMA, FNN, FC-LSTM) in single-dataset prediction and achieves performance comparable to state-of-the-art models like STGCN. Moreover, TL-GPSTGN excels in transfer learning scenarios, achieving high prediction accuracy on new road networks with minimal training data, showcasing its adaptability and efficiency.

\subsection{Future Work}
While TL-GPSTGN presents significant advancements in traffic prediction, several opportunities remain for further exploration and enhancement. First, integrating external factors such as weather conditions, social events, and traffic accidents could enrich the model's understanding of traffic dynamics. These external factors often play a crucial role in influencing traffic patterns but are not yet fully incorporated into the current framework.

Second, although graph pruning improves computational efficiency and migration performance, its impact on retaining critical structural and temporal features requires deeper investigation. Developing adaptive graph pruning strategies that dynamically adjust to varying network conditions and data quality could enhance robustness and generalizability.

Third, the current model focuses on transferring knowledge between road networks. Expanding its capabilities to multi-modal transportation systems, including pedestrian flows, public transit, and freight logistics, would significantly broaden its application in intelligent transportation systems. This would necessitate adapting the model to handle heterogeneous data sources and integrating multi-resolution spatial-temporal features.

Finally, while TL-GPSTGN achieves high accuracy with limited data, exploring self-supervised or unsupervised learning techniques could further reduce its dependence on labeled datasets. Leveraging abundant unlabeled traffic data could make the model more scalable and adaptable, particularly in regions where labeled data is sparse or unavailable.

In conclusion, this study advances the field of intelligent transportation systems by addressing critical challenges in traffic prediction, particularly in scenarios with limited historical data. By integrating transfer learning and graph pruning into a unified and robust framework, TL-GPSTGN establishes a new standard for predictive accuracy and adaptability across diverse and heterogeneous road networks. Beyond its technical contributions, the model holds significant potential for practical applications in areas such as smart city planning and autonomous vehicle systems. As intelligent transportation systems continue to evolve, the methodologies proposed in this work can serve as a foundation for developing more resilient, efficient, and adaptive urban mobility solutions.

\section*{Acknowledgment}

We sincerely thank Dr. Boyu Wang and Western University for their guidance and support throughout this project as part of our graduate course. The resources and knowledge provided in this course were instrumental in shaping our understanding and successful completion of this work.

\end{document}